\documentclass[hidelinks]{svjour3}

\usepackage{amsfonts,psfrag,layout,array,longtable,lscape,booktabs,dcolumn,amsmath,amssymb,colortbl,wasysym,threeparttable}
\usepackage[]{graphicx}
\usepackage{url}

\usepackage[numbers, square, comma, sort&compress]{natbib} 

\usepackage{subcaption}
\usepackage{microtype}

\usepackage{hyperref}

\RequirePackage{fix-cm}
\smartqed  

\usepackage{mathptmx}      
\usepackage[bottom]{footmisc}

\usepackage{footmisc}
\DefineFNsymbols{mySymbols}{{\ensuremath\dagger}{\ensuremath\ddagger}\S\P
	*{**}{\ensuremath{\dagger\dagger}}{\ensuremath{\ddagger\ddagger}}}
\setfnsymbol{mySymbols}

\usepackage{tabularx}
\newcolumntype{Y}{>{\raggedleft\arraybackslash}X}

\definecolor{Gray}{gray}{0.9}

\newcommand{\captionfonts}{\scriptsize}

\makeatletter  
\long\def\@makecaption#1#2{%
	\vskip\abovecaptionskip
	\sbox\@tempboxa{{\captionfonts #1: #2}}%
	\ifdim \wd\@tempboxa >\hsize
	{\captionfonts #1: #2\par}
	\else
	\hbox to\hsize{\hfil\box\@tempboxa\hfil}%
	\fi
	\vskip\belowcaptionskip}

\begin{document}
	
\title{Character-Based Handwritten Text Transcription with Attention Networks}


\author{Jason Poulos         \and
	Rafael Valle 
}

\institute{Jason Poulos \at
	Department of Statistical Science, Duke University, Durham, North Carolina, USA \\
	The Statistical and Applied Mathematical Sciences Institute, Durham, North Carolina, USA \\
	\email{jason.poulos@duke.edu}           
	\and
	Rafael Valle \at
	NVIDIA Corporation, Santa Clara, California, USA
}

\date{Received: date / Accepted: date}

\maketitle

\begin{abstract} 
The paper approaches the task of handwritten text recognition (HTR) with attentional encoder-decoder networks trained on sequences of characters, rather than words. We experiment on lines of text from popular handwriting datasets and compare different activation functions for the attention mechanism used for aligning image pixels and target characters. We find that softmax attention focuses heavily on individual characters, while sigmoid attention focuses on multiple characters at each step of the decoding. When the sequence alignment is one-to-one, softmax attention is able to learn a more precise alignment at each step of the decoding, whereas the alignment generated by sigmoid attention is much less precise. When a linear function is used to obtain attention weights, the model predicts a character by looking at the entire sequence of characters and performs poorly because it lacks a precise alignment between the source and target. Future research may explore HTR in natural scene images, since the model is capable of transcribing handwritten text without the need for producing segmentations or bounding boxes of text in images. 
\end{abstract}

\section{Introduction} \label{sec1}

Handwritten text recognition (HTR) on character sequences is an open research problem because it is harder to segment and recognize individual characters, rather words \cite{bluche2016}. Moreover, transcription models must solve the problem of finding and classifying characters at each time-step without knowing the alignment between the input sequence of image pixels and the target sequence of characters \cite{DBLP:journals/corr/LouradourK13}. 

Previous approaches to HTR include using a hidden Markov model (HMM), or HMM-neural network hybrid, to match image features to character labels. The HMM approach is outperformed by models that combine a single recurrent neural network (RNN) with a connectionist temporal classification (CTC) output layer \cite{graves2006connectionist, graves2009novel, liwicki2007, liwicki2012, wigington2017data,stuner2020handwriting}. The CTC-based models calculate a probability distribution over all possible target sequences, conditional on the input sequence. The CTC-based models assume strict monotonicity in input-target sequence alignments, and generally assume a target sequence length that is bounded by the input sequence length. 

In this work, we employ the encoder-decoder networks proposed by Deng et al. \cite{deng2016image}, which extends the encoder-decoder RNNs of Vinyals et al. \cite{vinyals2014grammar} and Bahdanau et al. \cite{bahdanau2014neural} for the problem of decompiling images into presentational markup. The encoder-decoder model encodes a variable-length sequence of characters into a fixed-length vector and then decodes the vector into a variable-length target label. Encoder-decoder RNNs are suitable for handling long sequences of data and have become standard for neural machine translation, speech recognition \cite{chorowski2015attention}, and image captioning \cite{xu2015show} tasks.

The model of Deng et al. consists of a convolutional neural network (CNN) that extracts visual features from the images and arrange the features on a grid. An RNN encoder re-encodes each row of the grid, learning additional features such as text directionality. Lastly, an RNN decoder outputs a character sequence one step at a time, using an attention mechanism to emphasize the most important columns of re-encoded features at each decoding step. The use of attention mechanism in the decoder relaxes the monotonicity assumption of the CTC-based model, and improves the ability of the encoder-decoder networks to learn the correct alignment between image pixels and target characters, and to extract the most relevant information for each part of the output sequence \cite{cho2014learning}. Attention-based networks are capable of modeling the language structures within the output sequence, rather than simply mapping the input to the correct output \cite{DBLP:journals/corr/ChoCB15}.

Encoder-decoder RNNs have been previously employed for recognizing text in natural images \cite{lee2016recursive, shi2016robust}, and more recently for HTR. Several recent papers propose a hybrid architecture consisting of a CNN to encode the input image and an RNN decoder to predict sequences of characters \cite{bluche2017gated,puigcerver2017multidimensional,chowdhury2018efficient,Zhang_2019_CVPR,kang2019candidate,kang2020unsupervised,xiao2020deep,retsinas2020wsrnet,belay2020learning}. For example, Sueiras et al. \cite{sueiras2018offline} and Kang et al. \cite{kang2018convolve} use attentional encoder-decoder networks very similar to ours, but train their model to transcribe words, rather than sequences of characters, and employ a word-based lexicon (i.e., a list of words found in the training set) for decoding. 

The main differentiator in our approach is that we employ a CNN to extract image features and a separate RNN encoder to re-encode the features so that the encoder can learn new features such as text directionality. Another difference is that we use an unidirectional RNN decoder to predict the sequence of characters. Gui et al. \cite{guiadaptive2018} train character-aware attention networks, but the architecture differs in that they use an attention-based bidirectional RNN decoder and CTC output layer to convert predictions made by the decoder into a character sequence. 

There are recent developments towards architecture based entirely on CNNs or attention mechanisms, bypassing any recurrence. Fully convolutional architectures have performed well against encoder-decoder networks on neural machine translation tasks \cite{gehring2017convs2s}, handwriting generation \cite{Fogel_2020_CVPR, davis2020text}, and HTR tasks \cite{poznanski2016cnn, such2018fully, coquenet2019have, ptucha2019intelligent,yousef2020accurate,Yousef_2020_CVPR}. The entirely attention-based transformer model initially proposed by Vaswani et al. \cite{vaswani2017attention} have outperformed encoder-decoder networks on several HTR tasks \cite{kang2020pay}.

In this work, we focus on developing character-aware models for HTR. Character-aware models view the input and output text lines as a sequence of characters rather than words, and each character prediction is explicitly conditioned on the previous character. These models are capable of making inferences about unseen source words and also generating unseen target words. In addition, character-aware models do not require lexicons because only characters are explicitly modeled \cite{DBLP:journals/corr/LingTDB15}. 

Our primary contributions are applying character-aware attention networks to the task of transcribing lines of unconstrained (i.e., cursive or overlapping) handwritten text and comparing different activation functions for the attention mechanism. Section \ref{imagetext} describes attention networks in the context of character-based HTR. Section \ref{experiment} describes the benchmark datasets used for the experiments and provides details on the network architecture and training. Section \ref{results} describes the results on benchmark datsets, comparing the performance of different attention mechanisms. Section \ref{conclusion} concludes and suggests directions for future research.

\section{Attention networks for character-based HTR} \label{imagetext}

The character-based HTR problem is one of converting images to hand-transcribed sequences of discrete characters. Following the notation of Deng et al., the input $\mathbf{x} \in \mathcal X$ is an image with height and width dimensions $\mathbb{R}^{H \times W}$. The target $\mathbf{y} \in \mathcal Y$ consists of a sequence of characters, $y_1, y_2, \ldots, y_T$, where $T$ is the sequence length and each character exists within a known vocabulary, $\Sigma$. The supervised task is to learn a function that maps $\mathcal X \rightarrow \mathcal Y$ using training example pairs ($\mathbf{x}$, $\mathbf{y}$). 

\begin{figure}[htbp] 
	\vskip 0.2in
	\begin{center}
		\includegraphics[width=\textwidth]{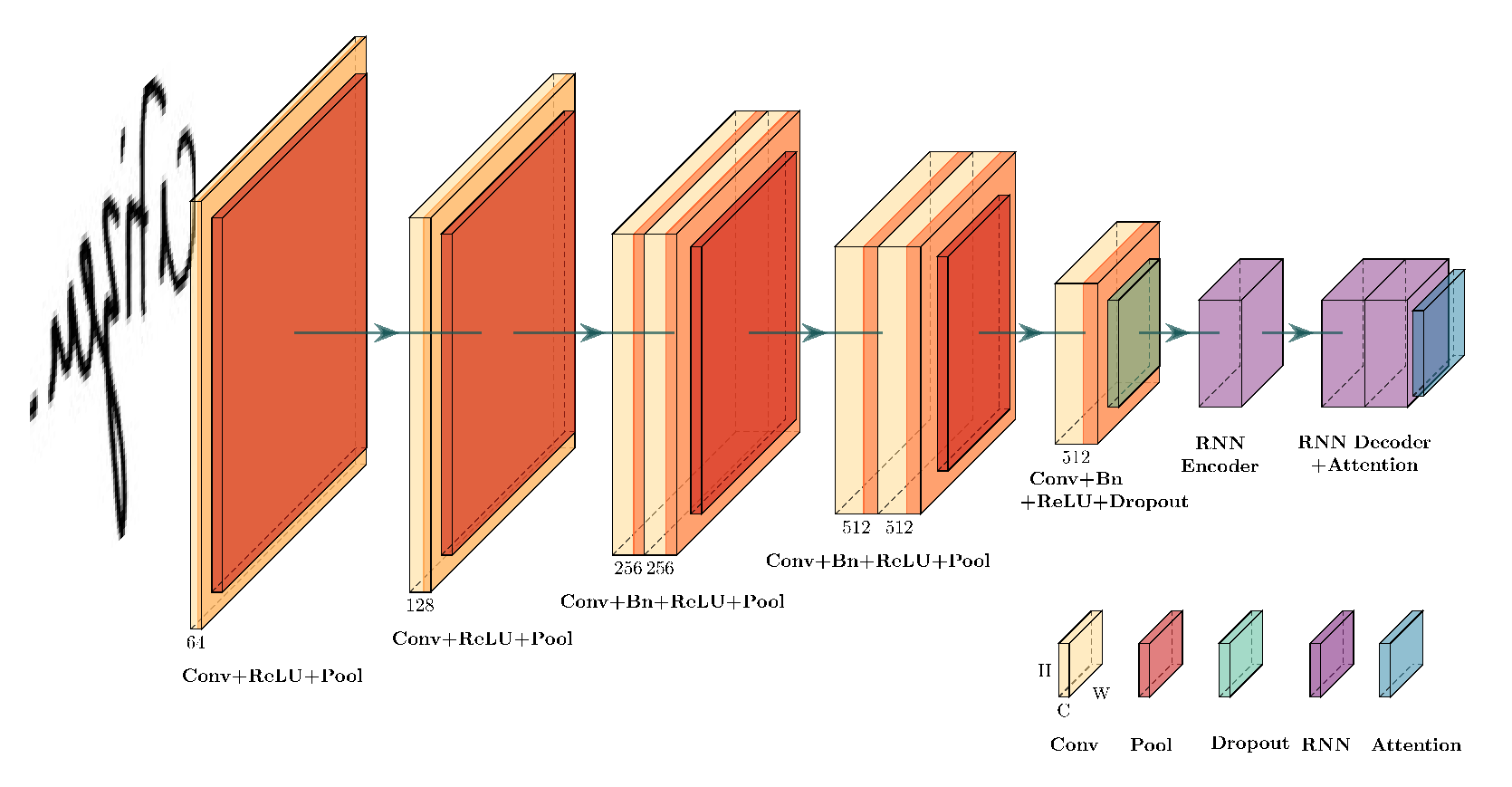}
	\end{center}
	\caption{Attention networks architecture. \emph{Notes:} `Conv': convolution layer; `Pool' max-pooling layer; `Bn': batch normalization.}
	\label{architecture}
	\vskip -0.2in
\end{figure}

The general architecture of the attention networks of Deng et al., which we extend for HTR, is illustrated in Fig.~\ref{architecture}. The CNN inputs $\mathbf{x}$ and arranges the visual features on a grid, $\mathbf{V}$ with dimensions $H' \times W' \times C$, where $C$ is the number of channels, and $H'$ and $W'$ are reduced dimensions following max pooling operations.

The RNN encoder slides across each row of $\mathbf{V}$, and at each time-step $t$, recursively updates a hidden state $\mathbf{h}_t$ using $\mathbf{v}_t \in \mathbf{V}$ as input:
\begin{equation} \label{encoder-hidden}
\mathbf{h}_t = f(\mathbf{v}_t, h_{t-1}; \theta),
\end{equation}
where $f(\cdot)$ is a nonlinear activation  and $\theta$ is a learned parameter. The encoder outputs a re-encoded feature grid $\mathbf{\tilde{V}}_{h, w} = \mathrm{RNN} (\mathbf{\tilde{V}}_{h, {w-1}}, \mathbf{V}_{h,w})$, for rows $h \in \{1, \ldots, H'\}$ and columns $w \in \{1, \ldots, W'\}$. Encoding row-wise is useful for transcription tasks because the encoder can learn sequential order information, such as text directionality. The networks capture column-wise sequential information by learning a positional encoding in the form of an initial hidden state, $\mathbf{\tilde{V}}_{h, 0}$, which is added to each row of $\mathbf{\tilde{V}}$.

The decoder RNN learns a conditional language model to give the probability of the next character given the history and re-encoded feature grid:
\begin{align}
p(y_{t+1} | y_{1},\ldots,y_{t}, \mathbf{\tilde{V}}) &=  \textrm{softmax} \left( \mathbf{W}_1 \textbf{o}_t \right), \label{language-model}\\
\text{where} \qquad \textbf{o}_t &= f(\textbf{W}_2 [\mathbf{h'}_t; \mathbf{c}_t]). \label{output}
\end{align}
In the above equations, the matrices $\mathbf{W}_1$ and $\mathbf{W}_2$ are learned parameters of the model, and the softmax activation function assigns probabilities over $\Sigma$. The hidden state of the RNN decoder, $\mathbf{h'}_t$, is updated recursively by
\begin{equation} \label{decoder-hidden}
	\mathbf{h'}_t = f \left( \mathbf{h'}_{t-1}, \mathbf{y}_{t-1}; \theta' \right),  
\end{equation}
where $\theta'$ is a learned parameter. The context vector, $\mathbf{c}_t$, provides the most important elements of the re-encoded feature grid at each $t$:
\begin{align}
\mathbf{c}_t &= \sum_{h,w} \boldsymbol{\alpha}_t \mathbf{\tilde{V}}_{h,w}, \label{context} \\
\text{where} \qquad \boldsymbol{\alpha}_t &= \text{softmax} (a (\mathbf{h'}_t, \mathbf{\tilde{V}}_{h,w})), \label{attention}\\ 
\text{and} \qquad a_{t,h,w} &= \boldsymbol{\beta}^{\top} f (\mathbf{W}_3 \mathbf{h'}_t + \mathbf{W}_4 \mathbf{\tilde{V}}_{h,w}),
\end{align}
where the vector $\boldsymbol{\beta}$ and matrices $\mathbf{W}_3$ and $\mathbf{W}_4$ are learned parameters, and the attention mechanism $a(\cdot)$ approximates the vector $\boldsymbol{\alpha}_t$ of unnormalized attention weights. 

The attention weights are distributed over columns of $\mathbf{\tilde{V}}_t$ so that each feature in the column is given identical weight, which is standard for typical character recognition tasks. This approach differs from the attention mechanism used by Deng et al., which places attention over rows and columns, so that attention weights vary for each element of $\mathbf{\tilde{V}}_t$, which may be more appropriate for complex images such as math formulas or tables. While the standard attention of Bahdanau et al. uses the softmax activation for Eq. \eqref{attention}, we experiment with two alternative activations to produce attention weights: sigmoid (i.e., Bernoulli) and linear (i.e., $a_t = e_t$). %

Finally, the networks are trained end-to-end to minimize the cross-entropy loss:
\begin{equation}\label{loss}
\mathcal{L}=\sum_{t=1}^{T}-\log p\left(y_{t+1} \mid y_{1}, \ldots, y_{t}, \mathbf{\tilde{V}}\right).
\end{equation}

\section{Experimental evaluation} \label{experiment}

We experiment on two widely-used HTR benchmark datasets, IAM (modern English) and RIMES (modern French), and two historical datasets, Saint Gall (9th c. Latin) and Parzival (13th c. German) \cite{marti2002iam,Grosicki2011,fischer2011transcription,fischer2009automatic}. The datasets, which are described in Table \ref{tab:benchmark-characteristics}, consist of images of handwritten text lines and their corresponding ground-truth transcriptions. 

\begin{figure}[htbp]
	\caption{Example preprocessed images from the benchmark datasets.\label{iam-rimes}}
	\vskip 0.15in
	\begin{center}
		\begin{subfigure}[b]{\linewidth}
			\includegraphics[width=\linewidth,height=1cm]{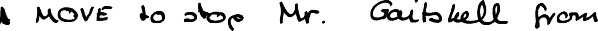}
			\caption{IAM}
		\end{subfigure}\\
		\begin{subfigure}[b]{\linewidth} 
			\includegraphics[width=\linewidth,height=1cm]{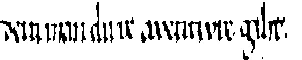} 
			\caption{Parzival}
		\end{subfigure}\\
		\begin{subfigure}[b]{\linewidth} 
			\includegraphics[width=\linewidth,height=1cm]{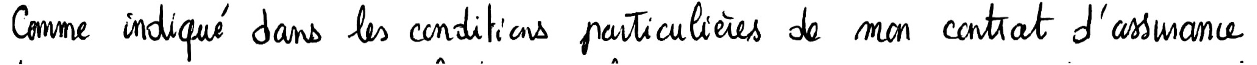} 
			\caption{RIMES}
		\end{subfigure}\\
		\begin{subfigure}[b]{\linewidth} 
			\includegraphics[width=\linewidth,height=1cm]{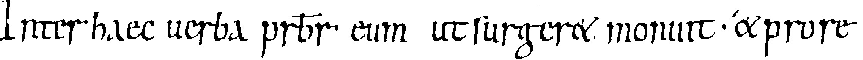} 
			\caption{Saint Gall}
		\end{subfigure}\\
	\end{center}
	\vskip -0.1in
\end{figure}

We follow the image preprocessing steps of Puigcerver et al. \cite{laia2016,puigcerver2017multidimensional}, which includes binarizing the images in a manner that preserves their original grayscale information \cite{villegas2015modification}, rescaling the images, and converting the images to JPEG format. Fig.~\ref{iam-rimes} provides an example of a preprocessed image from each benchmark dataset. 

\begin{table}[htbp]
	\centering
	\begin{threeparttable}
	\caption{Training, validation, and test set splits and language characteristics for benchmark datasets.}
	\label{tab:benchmark-characteristics}
	\vskip 0.15in
		\begin{tabular}{@{}lcccccccccc@{}}
			\toprule
			& \multicolumn{4}{c}{Lines}       & \multicolumn{3}{c}{Maximum length} & \multicolumn{3}{c}{Unique Characters} \\ \cmidrule(lr){2-5} 
			\cmidrule(lr){6-8} 
			\cmidrule(lr){9-11} 
			Dataset & Train  & Val.  & Test  & Total  & Train      & Val.      & Test      & Train       & Val.       & Test       \\
			\midrule
			IAM           & 6,161  & 966   & 2,915 & 10,042 & 81         & 73        & 95        & 79          & 76         & 75         \\
			Parzival      & 2,237  & 912   & 1,328 & 4,477  & 70         & 71        & 66        & 57          & 56         & 55         \\
			RIMES         & 10,171 & 1,162 & 778   & 12,111 & 100        & 110       & 94        & 97          & 88         & 85         \\ 
			Saint Gall    & 468    & 235   & 707   & 1,410  & 74         & 69        & 73        & 47          & 46         & 47         \\ \bottomrule
		\end{tabular}%
		\begin{tablenotes}
		\small
		\item \emph{Notes:} unique characters include case-sensitive alphanumeric characters, punctuation, and whitespace.
	\end{tablenotes}
\end{threeparttable}
\vskip -0.1in
\end{table}

\subsection{Evaluation} 

We measure the performance of the attention networks by comparing the estimated transcription $\mathbf{\hat{y}}$ with the ground-truth $\mathbf{y}$. Since the networks are trained on sequences of characters rather than words, we measure the Character Error Rate (CER) instead of the Word Error Rate. The CER is calculated as the edit distance normalized by the number of characters in the ground truth:
\begin{equation}\label{cer}
\text{CER} = \sum_t \frac{\text{Edit Distance}(y_t, \hat{y_t})}{| y_t |},
\end{equation} where the edit distance (or, Levenshtein distance), is the minimum number of insertions, substitutions, and deletions required to alter the target $y_t$ to the prediction $\hat{y_t}$ at each time-step. 
We also measure the character perplexity (CPPL) of the character-based conditional language model, which is the exponent of the cross-entropy loss defined in Eq. \eqref{loss}. Language models with smaller perplexity generally perform better in predicting characters given the history, and are thus strongly correlated with the CER \cite{wang2013new,bluche2015deep}.

\subsection{Implementation details}

When training the networks, we fix the image height to 64 pixels while maintaining the aspect ratio, group images with similar widths, and pad with whitespace to facilitate batching. We implement a biased importance sampling scheme to speed up training and decoding \cite{2014arXiv1412.2007J}.

The CNN converts the text line images into a sequence of visual feature vectors. It consists of seven convolutional layers, each followed by a Rectified Linear Unit (ReLU) activation and then a max-pooling layer to reduce the spatial size of the representation. The third, fifth, and seventh layers use batch normalization following the convolution in order to speed up training. Dropout is applied to the output of the seventh convolutional layer in order to prevent overfitting. Table \ref{cnn} provides further detail on the CNN specifications.

\begin{table}[htb]
	\centering
	\begin{threeparttable}
		\caption{CNN specification.\label{cnn}}
		\vskip 0.15in
		\begin{tabular}{lccccc}
			\hline 
			\multicolumn{4}{c}{Conv} & \multicolumn{2}{c}{Pool}	\\
			\hline 
			\multicolumn{1}{c}{$\#$ filters} & \multicolumn{1}{c}{Filter size} & \multicolumn{1}{c}{Stride size} & \multicolumn{1}{c}{Bn} & \multicolumn{1}{c}{Pool size} & \multicolumn{1}{c}{Stride size}	\\
			\hline 
			64&(3,3)&(1,1)& 	&(2,2)&(2,2) \\
			128&(3,3)&(1,1)&	&(2,2)&(2,2) \\
			256&(3,3)&(1,1)& \checkmark 	&-	&-	 \\
			256&(3,3)&(1,1)&	 & (2,1)& (2,1) \\
			512&(3,3)&(1,1)& \checkmark	& - & - \\
			512&(3,3)&(1,1)&	&	(2,1)&(2,1) \\
			512&(2,2)&(1,1)& \checkmark	& -& - \\
			\hline
		\end{tabular}
		\begin{tablenotes}
			\small
			\item \emph{Notes:} the sizes are ordered (height, width). See notes to Fig.~\ref{architecture}.
		\end{tablenotes}
	\end{threeparttable}
	\vskip -0.1in
\end{table}

Stacked on the CNN is a single-layer, bidirectional Long Short-Term Memory (BLSTM) encoder with 512 hidden units and a two-layer Gated Recurrent unit (GRU) decoder, each with 256 hidden units. The bidirectional recurrent layers allow the encoder to compute a representation that depends on both past and present characters in the sequence, and row-wise encoding refines the feature representation to include horizontal context. The attentional decoder interprets the feature representation, focusing on the most important columns of re-encoded features. 

We train the networks for 200 epochs with a batch size of 8, stochastic gradient descent to learn the parameter weights, and the Adam optimizer to adapt the learning rate. As a regularization strategy, we implement $\ell^2$ regularization loss and  data augmentation by applying random affine transformations to 20\% of the training set images, including scaling, translating, rotating, and shearing. In addition, we employ gradient norm clipping and gradient normalization in order to prevent exploding gradients. 

\section{Results} \label{results}

We train the attention networks without the assistance of any lexicon or explicit language model and record their performance in terms of CER and CPPL on the validation and test in Table \ref{eval}. The networks perform comparatively well on the Parzival and Saint Gall datasets, which have fewer training examples, and have shorter lines and vocabularies. The networks perform less well on the IAM and RIMES datasets, which have longer lines, and a larger vocabulary and number of training examples. 

\begin{table}[htb]
	\centering
	\begin{threeparttable}
		\caption{Attention networks: evaluation metrics on benchmark datasets.\label{eval}}
		\vskip 0.15in
		\begin{tabularx}{0.8\textwidth}{Xcccc}
			\toprule
			& \multicolumn{2}{c}{Val.}       & \multicolumn{2}{c}{Test}  \\ 
			\cmidrule(lr){2-3} 
			\cmidrule(lr){4-5}
			Dataset		 & CER (\%)  & CPPL  & CER (\%)  & CPPL    \\
			\midrule
			IAM           & 14.3   &  71,075.2  & 16.6   &  {\rm exp}(36.5)     \\
			Parzival      & 4.6    &  12.0     & 4.7    &   52.6    \\
			RIMES         & 11.1   &  811.9   & 12.1   &   92.4    \\
			Saint Gall    & 14.3   & 24.5      & 12.7   &   17,164.4   \\  \bottomrule
		\end{tabularx}%
		\begin{tablenotes}
			\small
			\item \emph{Notes:} networks trained with softmax attention. 
		\end{tablenotes}
	\end{threeparttable}
	\vskip -0.1in
\end{table}

Table \ref{benchmark-iam-rimes} compares the performance of the (softmax) attention networks on the IAM and RIMES test set with models in the existing literature. The attention networks achieve a CER of 16.6\% on the IAM dataset, which outperforms CTC models that encode image features using LSTMs or multidimensional LSTMs (MDLSTMs) \cite{graves2009offline}, but does not approach the current state-of-the-art model of Bluche and Messina \cite{bluche2017gated}, which combines convolutional and recurrent layers for encoding with a CTC decoder.

A direct comparison against most of the models in the existing literature is not possible because most of the existing models rely on domain-specific lexicons, and explicit language models for decoding. Bluche \cite{bluche2015deep}, for example, uses a word-based lexicon and a word-based language model. The model of Bluche \cite{bluche2016}, which combines a MDLSTM encoder and a softmax attention-enhanced bidirectional LSTM decoder, inputs and outputs at the character-level, although the decoder output is not conditioned on the previous character. The aforementioned model is also trained with curriculum learning and with a slightly larger training set. The state-of-the-art model of Bluche and Messina \cite{bluche2017gated}, in comparison, uses a hybrid word and character-based language model. Gui et al. \cite{guiadaptive2018} also train character-aware attention networks, but with a CTC output layer to perform the transcription. Michael et al. \cite{michael2019evaluating} is the most comparable to our work because the authors train character-aware attention networks without the use of a language model. 

\begin{table}[t]
	\centering
	\begin{threeparttable}
	\caption{Benchmark comparison: test set CER on IAM and RIMES datasets.\label{benchmark-iam-rimes}}
	\vskip 0.15in
		\begin{tabular}{llcccc}
		\toprule
		Model & Source & LM  & CB & IAM CER (\%) & RIMES CER (\%) \\
		\midrule
		CNN + BLSTM + CTC & \cite{bluche2017gated}  & \checkmark   &  & 3.2 & 1.9 \\ 
		MDLSTM + CTC & \cite{voigtlaender2016handwriting}  & \checkmark   &  & 3.5 & 2.8   \\ 
		MDLSTM + MLP/HMM & \cite{castro2018boosting}  & \checkmark   &  & 3.6 & - \\ 
		MDLSTM + CTC & \cite{bluche2015deep}  & \checkmark   &  & 4.4 & 3.5 \\ 
		CNN + LSTM + CTC & \cite{puigcerver2017multidimensional}  & \checkmark &  & 4.4 & 2.3 \\ 
		MDLSTM + Attention & \cite{bluche2016joint} & \checkmark  &   & 4.4 & 3.5 \\   
		Transformer & \cite{kang2020pay}  & &   & 4.6 & - \\  
		LSTM + HMM & \cite{doetsch2014fast}  & \checkmark &   & 4.7 & 4.3  \\  
		LSTM + HMM & \cite{voigtlaender2015sequence}  & \checkmark &   & 4.8 & 4.3 \\ 
		CNN + LSTM + Attention & \cite{michael2019evaluating}  & \checkmark & \checkmark &   4.8 & -\\ 
		CNN + CTC & \cite{yousef2020accurate}  &  & \checkmark &   4.9 & -\\  
		CNN + LSTM + Attention & \cite{coquenet2020endtoend}  &  & \checkmark &   4.9 & -\\  
		LSTM + HMM  & \cite{kozielski2013improvements} & \checkmark   &  & 5.1 & 4.6 \\ 
		MDLSTM + CTC & \cite{pham2014dropout}  & \checkmark &   & 5.1 & 3.3\\  
		CNN + BLSTM + Attention + CTC	& \cite{guiadaptive2018}  &    &  & 5.1 & -\\ 
		CNN + BLSTM & \cite{dutta2018improving}  &   &  & 5.7 & 5.0 \\ 
		CNN + BGRU + GRU + Attention & \cite{kang2019candidate}  & \checkmark   &		 &   5.7  &   2.6\\ 
		CNN + CTC & \cite{Huang2020}  &  &  &   6.1 & 3.4\\\ 
		MDLSTM + CTC & \cite{bluche2016}  & \checkmark &  &   6.6 & -\\ 
		CNN + BGRU + GRU & \cite{kang2018convolve}  &   &		 &   6.8 & - \\ 
		CNN + BLSTM + LSTM	& \cite{chowdhury2018efficient}  &  &  & 8.1 & 3.5\\  
		GMM/HMM & \cite{kozielski2013open}  & \checkmark  &  & 8.2 & -\\ 
		CNN + LSTM + Attention	& \cite{sueiras2018offline}  &  &  &   8.8 & -\\ 
		CNN + LSTM + CTC	& \cite{krishnan2018word}  &  &  &   9.7 & - \\
		MLP/HMM  & \cite{5551147}  &  \checkmark &  &   9.8 & -\\  
		MDLSTM + CTC & \cite{chen2017simultaneous}  &  \checkmark &  &   11.1 &   8.29 \\  
		MLP/HMM & \cite{dreuw2011hierarchical}  &  \checkmark &  &   12.4 & -\\ 
\rowcolor{Gray}		CNN + BLSTM + GRU + Attention & Ours  &  &  \checkmark &      16.6 & 12.1 \\ 
		MDLSTM + CTC & \cite{DBLP:journals/corr/LouradourK13}  &  &   &   17.0 & - \\ 
		BLSTM + CTC & \cite{liwicki2012}  & \checkmark &  &   18.2 & - \\
		CNN + LSTM + Attention & \cite{coquenet2020endtoend}  &  & \checkmark &  - & 3.1 \\  
		CNN + BLSTM + Attention & \cite{doetsch2016bidirectional}  & \checkmark   & & - & 5.8 \\ 
		HMM/MLP & \cite{menasri2012a2ia}  &  \checkmark &  & - &  7.2\\ 
		BLSTM + CTC & \cite{soullard2019ctcmodel}  &   &  &  - & 7.6\\ 
		\bottomrule
	\end{tabular}

	\begin{tablenotes}
	\item \emph{Notes:} `BGRU': bi-directional GRU; `CB': model is character-based; `GMM': Gaussian mixture model; `LM': explicit language model used for decoding; `MLP': multilayer perceptron.
	\end{tablenotes}
\end{threeparttable}
\vskip -0.1in
\end{table}

\subsection{Comparing attention distributions}

\begin{figure}[htbp] 
	\vskip 0.2in
	\begin{subfigure}[b]{0.33\linewidth}
		\begin{center}
			\includegraphics[width=\linewidth]{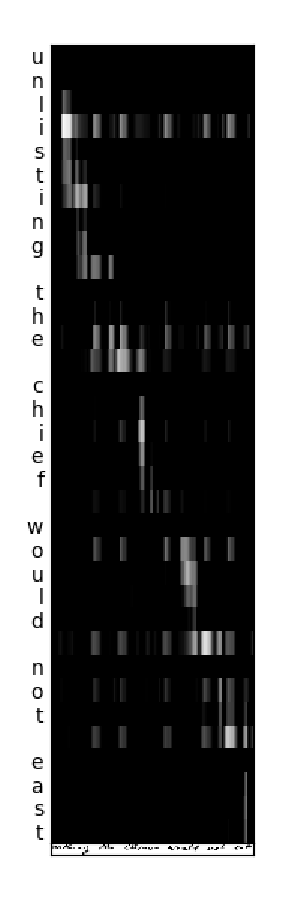}
			\caption{Linear}
		\end{center}
	\end{subfigure}%
	\begin{subfigure}[b]{0.33\linewidth}
		\begin{center}
			\includegraphics[width=\linewidth]{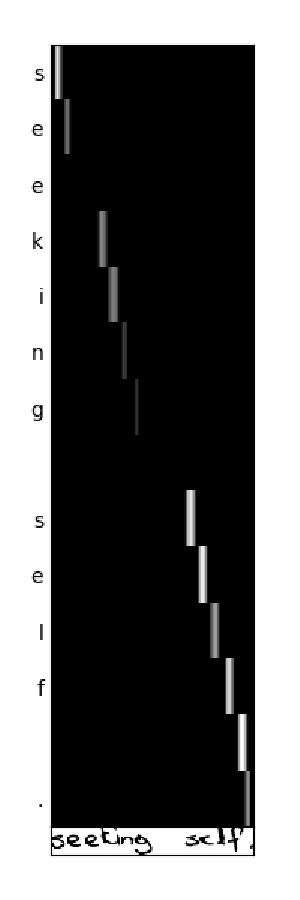}
			\caption{Sigmoid}
		\end{center}
	\end{subfigure}%
	\begin{subfigure}[b]{0.33\linewidth}
		\begin{center}
			\includegraphics[width=\linewidth]{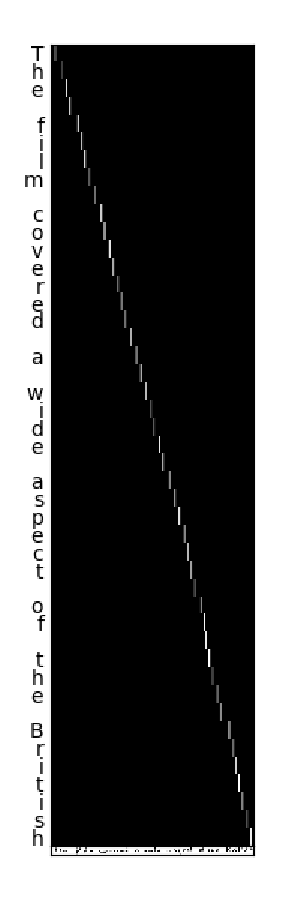}
			\caption{Softmax}
		\end{center}
	\end{subfigure}%
	\caption{Visualization of the source attention distribution over the input image (horizontal axis). The vertical axis is the transcription. Each row traces the attention weights over the source line at each step of decoding, in grayscale (0: black, 1: white).}
	\label{attention-dist}
	\vskip -0.2in
\end{figure}

In order to gain insight into how the attention mechanism learns alignment between the source and target character, we plot in Fig.~\ref{attention-dist} a visualization of the source attention distribution for attention networks trained on the IAM dataset. Each row traces the attention weights over the source line at each step of decoding. White values reflect intensity of attention while absence of attention is black. 

Softmax attention predicts a character by focusing heavily on single characters, whereas the attention distribution for sigmoid focus on multiple characters at each time-step. Softmax attention is able to learn a linear alignment whereas the alignment generated by sigmoid attention is linear and less precise.\footnote{Similarly, Kim et al. \cite{DBLP:journals/corr/KimDHR17} find that softmax attention performs better than sigmoid attention on word-to-word machine translation tasks.} When a linear function is used to obtain the attention weights, the model predicts a character by looking at the entire sequence of characters, and there is no clear structure in the alignment. 

\begin{figure}[ht] 
	\vskip 0.2in
	\captionsetup[subfigure]{justification=centering}
	\begin{subfigure}[]{\linewidth}
		\begin{center}
			\includegraphics[width=\linewidth]{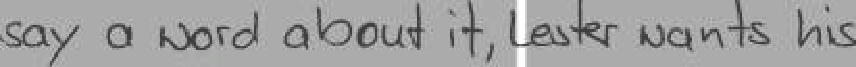} 
			\caption{Actual: \emph{say a word about it , Lester wants his}; \\
				Predicted: \emph{say a word about it , \underline{l}ester wants his}}
		\end{center}
	\end{subfigure}
	\begin{subfigure}[]{\linewidth}
		\begin{center}
			\includegraphics[width=\linewidth]{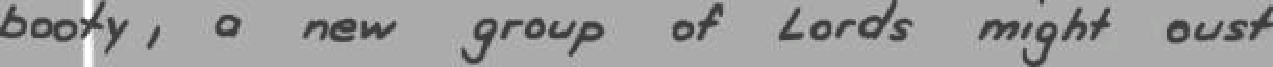} 
			\caption{Actual: \emph{booty , a new group of Lords might oust}; \\
				Predicted: \emph{boo\underline{k}y , o new group of Lords might oust}}
		\end{center}
	\end{subfigure}
	\begin{subfigure}[]{\linewidth}
		\begin{center}
			\includegraphics[width=\linewidth]{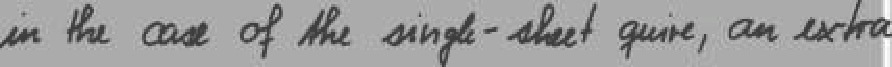} 
			\caption{Actual: \emph{in the case of the single-sheet quire , an extra}; \\
				Predicted: \emph{in the case of the single-sheet quire , an extra\underline{a}}}
		\end{center}
	\end{subfigure}
	\begin{subfigure}[]{\linewidth}
		\begin{center}
			\includegraphics[width=\linewidth]{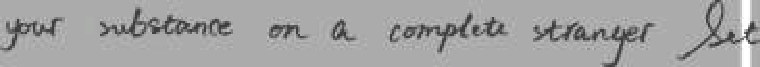} 
			\caption{Actual: \emph{your substance on a complete stranger . Set}; \\
				Predicted: \emph{your subteance on a complete stranger , \underline{f}ut}}
		\end{center}
	\end{subfigure}
	\caption{Incorrect IAM transcriptions and visualized softmax attention. White lines indicates the attended regions and underlines in the transcription indicate the corresponding character.}
	\label{attention-incorrect}
	\vskip -0.2in
\end{figure}

\begin{figure}[ht]
	\vskip 0.2in
	\captionsetup[subfigure]{justification=centering}
	\begin{subfigure}[]{\linewidth}
		\begin{center}
			\includegraphics[width=\linewidth]{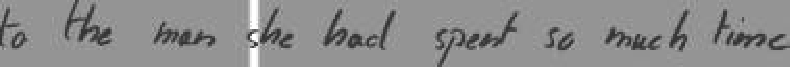} 
			\caption{Actual/predicted: \emph{to the man \underline{s}he had spent so much time}}
		\end{center}
	\end{subfigure}
	\begin{subfigure}[]{\linewidth}
		\begin{center}
			\includegraphics[width=\linewidth]{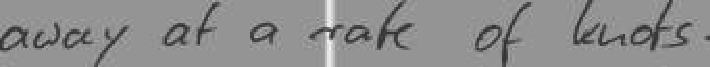} 
			\caption{Actual/predicted: \emph{away at a \underline{r}ate of knots .}}
		\end{center}
	\end{subfigure}            
	\begin{subfigure}[]{\linewidth}
		\begin{center}
			\includegraphics[width=\linewidth]{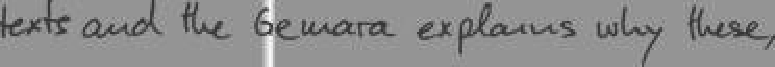} 
			\caption{Actual/predicted: \emph{texts and the \underline{G}emara explains why these ,}}
		\end{center}
	\end{subfigure}
	\begin{subfigure}[]{\linewidth}
		\begin{center}
			\includegraphics[width=\linewidth]{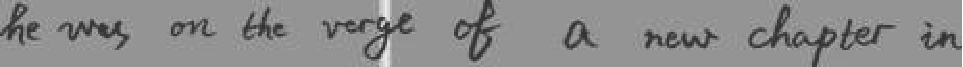} 
			\caption{Actual/predicted: \emph{he was on the ver\underline{g}e of a new chapter in}}
		\end{center}
	\end{subfigure}
	\caption{Correct IAM transcriptions and visualized softmax attention. See footnotes to Fig. \ref{attention-incorrect}.}
	\label{attention-correct}
	\vskip -0.2in
\end{figure}

In order to determine how the model makes mistakes, we visualize attention on the input image drawn from the IAM dataset. For example, the model tends to produce errors when characters are skewed (Fig.~\ref{attention-incorrect} [b]), have long tails (Fig.~\ref{attention-incorrect} [a] and [c]), or written in uppercase cursive (Fig.~\ref{attention-incorrect} [d]). Fig.~\ref{attention-correct}, which provides examples of correct IAM transcriptions and visualized softmax attention, shows that the model can correctly predict illegible handwriting (Fig.~\ref{attention-correct} [b]) because it leverages information from the entire input sequence. 

\section{Conclusion and future directions} \label{conclusion}

The paper approaches the task of handwritten text transcription with attention-based encoder-decoder networks trained to handle sequences of characters rather than words. The attention networks are domain and language-agnostic because they are trained without the aid of a lexicon or explicit language model. 

We train the model on lines of text from a popular handwriting dataset and experiment with different activation functions for the attention mechanism. Our results show that softmax attention focuses heavily on individual characters, while sigmoid attention focuses on multiple characters at each step of the decoding. When the sequence alignment is one-to-one, softmax attention is able to learn a more precise alignment at each step of the decoding whereas the alignment generated by sigmoid attention is much less precise. When the model has linear attention, the model predicts a character by looking at the entire sequence of characters and performs poorly because it lacks a precise alignment between the source and text output. 

Our primary contributions are applying character-aware attention networks to the task of handwritten text line transcription and also comparing attention configurations for the decoder. Future work might apply attention networks to the problem of HTR in natural scene images \cite{veit2016coco}. Previous literature has focused on recognizing printed text in natural scene images using standard methods in computer vision for segmentation \cite{jaderberg2016reading}. The attention networks used in this paper are capable of transcribing handwritten text without the need for producing segmentations or bounding boxes of text in images, so the model can potentially transcribe handwritten text in natural scene images without preprocessing. 
%

\section*{Declarations}

\textbf{Funding} Poulos acknowledges support of the National Science Foundation Graduate Research Fellowship under Grant DGE-1106400, and the National Science Foundation under Grant DMS-1638521 to the Statistical and Applied Mathematical Sciences Institute. Any opinions, findings, and conclusions or recommendations expressed in this material are those of the authors and do not necessarily reflect the views of the National Science Foundation. \\

\noindent\textbf{Availability of data and material} The IAM, Saint Gall, and Parzival datasets can be downloaded from:\\
 \url{https://fki.tic.heia-fr.ch/databases}. \\ \\
 \noindent
The RIMES dataset can be downloaded from:\\
\url{http://www.a2ialab.com/doku.php?id=rimes_database:start}. \\

\noindent\textbf{Code availability} Implementation code is available at the repository:\\
 \url{https://github.com/jvpoulos/Attention-OCR/}.\\

\noindent\textbf{Conflicts of interest} The authors declare no conflicts of interest.

\clearpage
\bibliography{refs}
\bibliographystyle{spbasic_unsort}

\end{document}